\documentclass[10pt,conference]{IEEEtran}
\IEEEoverridecommandlockouts
\usepackage{balance}
\usepackage{float}
\usepackage{cite}
\usepackage{amsmath,amssymb,amsfonts}
\usepackage{textcomp}
\usepackage{graphicx}
\usepackage{epstopdf}
\usepackage{xcolor}
\usepackage{color}
\usepackage{amsmath}
\usepackage[linesnumbered,ruled,vlined]{algorithm2e}
\usepackage{hyperref}
\usepackage{cleveref}
\hypersetup{hidelinks}
\usepackage{multirow}
\usepackage{bm}
\usepackage[normalem]{ulem}

\usepackage{tikz}
\usepackage{pgfplots}
\usepackage{microtype}
\usepackage{graphicx}
\usepackage{subcaption} %
\usepackage{booktabs} %
\usepackage{adjustbox}
\usepackage{xspace}
\usepackage{multirow}
\usepackage{thmtools, thm-restate} %

\ifCLASSOPTIONcompsoc
\else
\fi

\def\BibTeX{{\rm B\kern-.05em{\sc i\kern-.025em b}\kern-.08em
    T\kern-.1667em\lower.7ex\hbox{E}\kern-.125emX}}

\begin{document}
\title{
Continual Learning with Diffusion-based Generative Replay for Industrial Streaming Data
}

\author{\IEEEauthorblockN{
        Jiayi He\IEEEauthorrefmark{1},
        Jiao Chen\IEEEauthorrefmark{1},
        Qianmiao Liu\IEEEauthorrefmark{1},
        Suyan Dai\IEEEauthorrefmark{1},
        Jianhua Tang\IEEEauthorrefmark{1}\IEEEauthorrefmark{2},
        and Dongpo Liu\IEEEauthorrefmark{1}\IEEEauthorrefmark{3}
        }

    \IEEEauthorblockA{\IEEEauthorrefmark{1} Shien-Ming Wu School of Intelligent Engineering, South China University of Technology, Guangzhou, China}
    \IEEEauthorblockA{\IEEEauthorrefmark{2} Pazhou Lab, Guangzhou, China}
    \IEEEauthorblockA{\IEEEauthorrefmark{3} China Academy of Information and Communications Technology, Beijing, China}
        \IEEEauthorblockA{ \{202164020171, 202110190459, 202030010312, 202066200091\}@mail.scut.edu.cn,
        jtang4@e.ntu.edu.sg, liudongpo@caict.ac.cn}

\thanks{Pre-print, manuscript submitted to IEEE. The corresponding authors are Jianhua Tang and Dongspo Liu. The work was supported in part by  \textquotedblleft Climbing Plan\textquotedblright \ Guangdong University Students Science and Technology Innovation Cultivation Special Funds Subsidized Projects under Grant pdjh2023b0040.
}
}

\maketitle

\begin{abstract}
The Industrial Internet of Things (IIoT) integrates interconnected sensors and devices to support industrial applications, but its dynamic environments pose challenges related to data drift.
Considering the limited resources and the need to effectively adapt models to new data distributions,
this paper introduces a Continual Learning (CL) approach, \textit{i.e.}, Distillation-based Self-Guidance (DSG), to address challenges presented by industrial streaming data via a novel generative replay mechanism.
DSG utilizes knowledge distillation to transfer knowledge from the previous diffusion-based generator to the updated one, improving both the stability of the generator and the quality of reproduced data, thereby enhancing the mitigation of catastrophic forgetting.
Experimental results on CWRU, DSA, and WISDM datasets demonstrate the effectiveness of DSG.
DSG outperforms the state-of-the-art baseline in accuracy, demonstrating improvements ranging from 2.9\% to 5.0\% on key datasets, showcasing its potential for practical industrial applications.
\end{abstract}


\section{Introduction}

The rapid advancement of the Industrial Internet of Things (IIoT) has led to the integration of thousands of interconnected sensors and computing devices \cite{10001430}.
These systems support various industrial artificial intelligence (AI) applications, including industrial anomaly detection, mechanical predictive maintenance, and industrial human action recognition, showcasing the immense potential of IIoT in improving production efficiency and ensuring equipment operation safety.

In real-world industrial settings, IIoT devices frequently function under dynamically changing conditions. For instance, data gathered by autonomous guided vehicles and robots may experience deviations stemming from shifts in location, and visual data obtained by surveillance cameras could be influenced by fluctuations in lighting and seasonal changes. This phenomenon, known as \textit{data drift}, can significantly impair the intelligent perception capabilities of devices and consequently degrade the performance of associated applications.
Traditional machine learning methods often require retraining models to adjust to new conditions resulting from data drift.
However, this approach often leads to forgetting previously acquired knowledge, known as \textit{catastrophic forgetting} \cite{kirkpatrick2017overcoming}.
This outcome escalates the computational cost of model retraining and wastes computational resources.
Therefore, enabling models to overcome catastrophic forgetting, caused by data drift, becomes crucial while preserving knowledge from previous data.

\begin{figure}[t]
    \centering
    \includegraphics[width=0.95\linewidth]{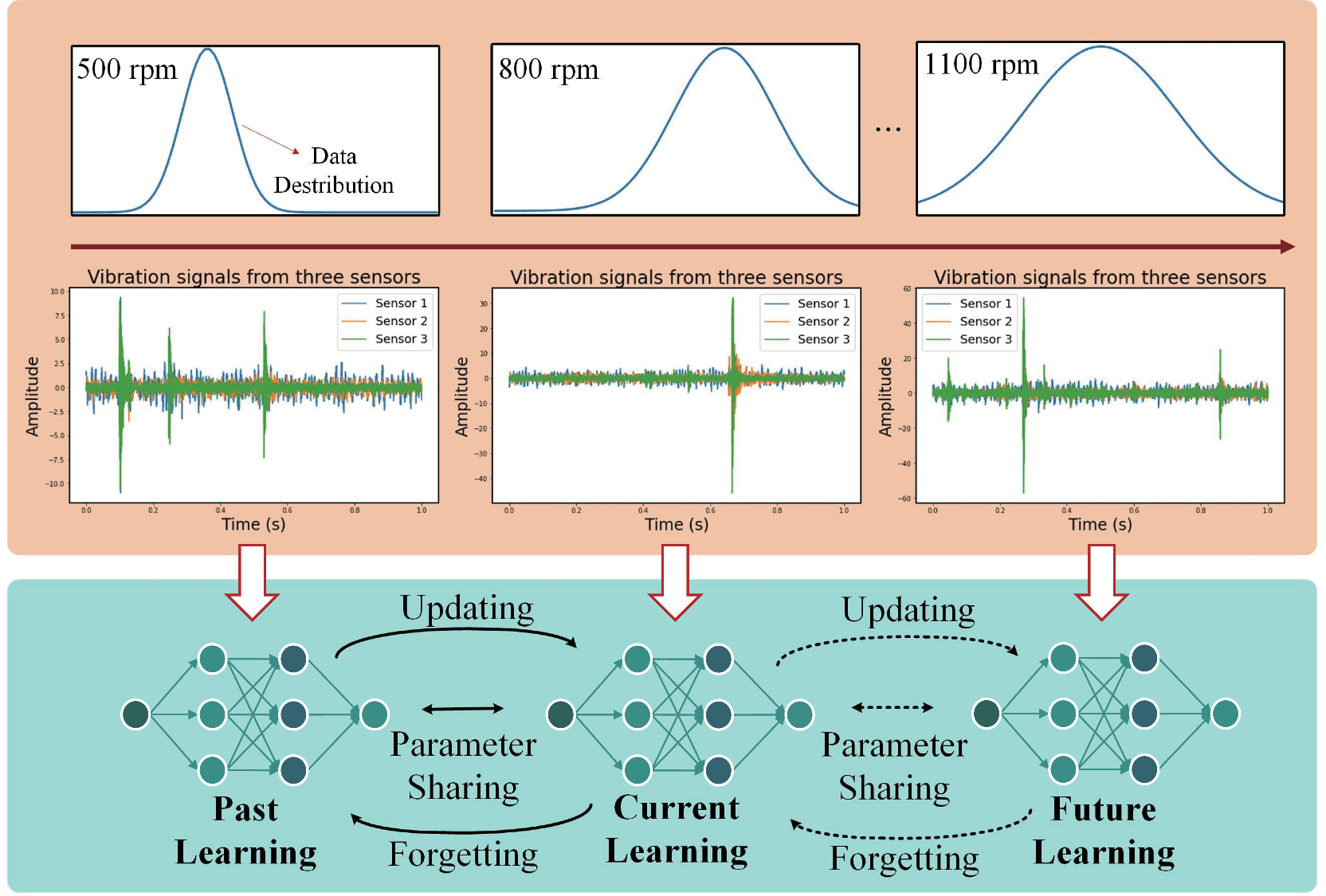}
    \caption{Continual learning of the bearing fault diagnosis model. With streaming data from different operating conditions, the model is required to be updated to cope with current data while mitigating the forgetting of previous learning.
    }
    \label{fig:CL}
\end{figure}

Continual Learning (CL),  also referred to as Incremental Learning or Lifelong Learning, aims to learn a sequence of data/tasks continuously without forgetting previously acquired knowledge in non-stationary settings (Fig.~\ref{fig:CL}).
Among various methods of CL, replay-based methods replay a portion of data from previous tasks to retain prior knowledge while learning new tasks.
Generative-based replay methods involve training an extra model to produce pseudo-data for replay. They are particularly suitable for IIoT applications with limited storage and computational resources, as they do not require a memory buffer to store previous data.
Recently, advanced generators such as Deep Diffusion Probabilistic Models have gained popularity for data generation tasks, exhibiting outstanding performance in various domains \cite{ho2020denoising,song2019generative,song2020score}.
Some generative-based replay CL methods employ diffusion probabilistic models as their generators, achieving an increase in the effectiveness of the generated data \cite{kim2024sddgr}.

Traditional generative replay methods have a limitation in that they use replayed data to update the generator \cite{shin2017continual}, leading to lower-quality data for generating earlier tasks and gradual deviation of generated samples from the distributions of those tasks \cite{gao2023ddgr}.
To preserve the quality of replayed data while ensuring that new data generation does not deviate from the original data distributions,
this paper introduces the Distillation-based Self-Guidance (DSG) method, which employs knowledge distillation techniques to transfer knowledge between old and new generators.
In DSG, the new generator trained for a new task acts as a student, while its preceding generator functions as a teacher, retaining knowledge from previous tasks.
As the student learns input from a new task, guidance is provided through distillation loss from the teacher, facilitating the transfer of knowledge from previous task inputs and enabling guidance between the old and new generators (see Fig.~\ref{fig:distillation}).

Our main contributions are:

$\bullet$ We develop a CL method tailored for the IIoT to address the unique challenges posed by evolving industrial streaming data. This method integrates a novel generative replay mechanism suitable for deployment on resource-constrained IoT devices.
$\bullet$ We introduce the Distillation-based Self-Guided (DSG) approach, which leverages a diffusion probabilistic model to generate previous data.
DSG facilitates the effective transfer of knowledge from previous tasks to the new generator, thereby enhancing the model's capability to mitigate catastrophic forgetting.

$\bullet$ We conduct comprehensive experiments and evaluations on the CWRU, DSA, and WISDM datasets to validate the effectiveness of our proposed methods. Our results demonstrate the enhanced model performance and practical utility of our method in industrial applications.

\section{Related Works}

Our work focuses on utilizing the distillation-guided diffusion probabilistic model for synthesizing data in the replay phase of CL.
In this section, we provide a brief overview of the background related to the framework, the replay generator, and the guidance for updating the generator.

\subsection{Continual Learning}
CL progressively learns a sequence of tasks without forgetting previously acquired knowledge in dynamic scenarios.
Regularization-based methods apply regularization terms or knowledge distillation to constrain model updates \cite{kirkpatrick2017overcoming}.
Replay-based methods learn old and new data when adapting to new tasks, with the old data obtained through experience replay \cite{rolnick2019experience} or generative replay \cite{shin2017continual}.
Parameter isolation methods train specific parameters in network for each task \cite{mallya2018packnet}.

\subsection{Diffusion Probabilistic Model}
Diffusion Probabilistic Models are deep generative models that have two diffusion processes based on a Markov chain: (1) the forward diffusion process, in which Gaussian noise is gradually added to the input data, and (2) the reverse diffusion process, wherein a generative model progressively learns to reverse the diffusion process to recover the original input from the noise.
Three standard foundations of diffusion probabilistic models include Denoising Diffusion Probabilistic Models (DDPMs) \cite{ho2020denoising}, Noise Conditioned Score Networks (NCSNs) \cite{song2019generative}, and Stochastic Differential Equations (SDEs) \cite{song2020score}.
We utilized DDPMs in this paper.
DDPMs \cite{ho2020denoising} trains a sequence of diffusion probabilistic models to reverse noise, learning the reverse process by estimating the noise at each step.

\subsection{Knowledge Distillation}
Knowledge distillation facilitates knowledge transfer from a cumbersome model to a smaller one.
Offline distillation methods typically involve training a large and complex teacher model to perform well on a specific task, and training a simpler student model to closely imitate the output of the teacher model on the same task \cite{hinton2015distilling}.
Online distillation methods simultaneously update the teacher and the student and employ end-to-end training \cite{zhang2018deep}.
Self-distillation methods distill knowledge internally, where the student and teacher originate from the same networks \cite{zhang2019your}.
Our DSG is categorized as an offline distillation method.

\section{Methodology}
\subsection{Problem Definition}
CL aims to train a neural network \( M_{\bm{\theta}} \), where \( \bm{\theta} \) represents the set of parameters of the model, to learn from a sequence of $N$ tasks, denoted as \( \mathcal{T} = \{  \mathcal{T}_{1}, \cdots, \mathcal{T}_n, \cdots, \mathcal{T}_{N} \} \).
Each task $\mathcal{T}_n=\{(\textbf{x}_{n,i},y_{n,i})\}_{i=1}^{|\mathcal{T}_n|}$ is a supervised dataset, consisting of the pair of input $\textbf{x}_{n,i}$ and its corresponding label $y_{n,i}$.
The primary challenge of CL lies in enabling the network to acquire current knowledge from current task $\mathcal{T}_{n}$ without losing previously learned information from $\left \{  \mathcal{T}_{1}, \cdots,  \mathcal{T}_{n-1}\right \}$.

Generative replay, as a method in CL, aims to minimize both the loss function related to the current task $\mathcal{T}_n$, denoted as $\mathcal{L}_{task}(\cdot,\cdot)$, and the replay loss on synthetically generated data from previous tasks, denoted as $\mathcal{L}_{replay}(\cdot,\cdot)$, concurrently.
The compound loss is given by:

\begin{equation}
\min_{\bm{\theta}} \left[ \mathcal{L}_{task}(M_{\bm{\theta}}, \mathcal{T}_n) + \mathcal{L}_{replay}(M_{\bm{\theta}}, \hat{\mathcal{T}}_{1:n-1}) \right],
\end{equation}
where $\hat{\mathcal{T}}_{1:n-1}$ denotes the reconstructed datasets from task $\mathcal{T}_{1}$ to $\mathcal{T}_{n-1}$.
High-quality generated data that closely approximates the original distribution can result in reduced $\mathcal{L}_{replay}$ during the learning of new information. Consequently, as the model tackles new tasks, it retains knowledge from previous tasks more effectively, thereby mitigating catastrophic forgetting.

\subsection{Diffusion-based Generative Replay}
We use a DDPM as the generator $G_{\bm{\phi}}$.
We begin by introducing the forward process and the reverse process of the DDPM, followed by the training objectives and sampling methods.
For clarity, in this section, we omit the notations of task index $n$ and sample index $i$, focusing solely on a data sample $\textbf{x}_{0}$ from the real data distribution $q(\textbf{x})$ as an example.

\subsubsection{Forward Diffusion Process}

The original data $\textbf{x}_{0}$ undergoes a gradual transformation into noise data $\textbf{x}_{T}$ through a series of $T$ ($0 \leq \tau \leq T$) steps, which is accomplished by gradually adding Gaussian noise.
The set $\{\beta_\tau\in(0,1)\}_{\tau=1}^{T}$ represents variance schedules determining step sizes. $\mathbf{I}$ denotes the identity matrix, and $\mathcal{N}$ denotes the Gaussian distribution. $\mathcal{N}(\textbf{x}_\tau;\sqrt{1-\beta_\tau}\textbf{x}_{\tau-1},\beta_\tau\mathbf{I})$ specifies the Gaussian distribution of the random variable $\textbf{x}_\tau$ given $\textbf{x}_{\tau-1}$, where $\sqrt{1-\beta_\tau}\textbf{x}_{\tau-1}$ and $\beta_\tau\mathbf{I}$ represent the mean and variance of the distribution, respectively.
The forward process from $\textbf{x}_{\tau-1}$ to $\textbf{x}_{\tau}$ is described by
\begin{equation}
    q(\textbf{x}_\tau|\textbf{x}_{\tau-1})=\mathcal{N}(\textbf{x}_\tau;\sqrt{1-\beta_\tau}\textbf{x}_{\tau-1},\beta_\tau\mathbf{I}).
\end{equation}
The distribution of the sample $\textbf{x}_{1:T}$ can be obtained by the product of
\begin{equation}
q(\textbf{x}_{1:T}|\textbf{x}_0)=\prod_{\tau=1}^{T}q(\textbf{x}_\tau|\textbf{x}_{\tau-1}),
\end{equation}
where $\textbf{x}_{T}$ is pure Gaussian noise.

Let $\alpha_\tau=1-\beta_\tau$, $\bar{\alpha}_\tau=\prod_{i=1}^\tau\alpha_i$, and noise $\bm{\epsilon}_\tau\sim\mathcal{N}(\mathbf{0},\mathbf{I})$, $\textbf{x}_{\tau}$ can be expressed by the original input $\textbf{x}_{0}$ through
\begin{equation}
   \textbf{x}_{\tau} =\sqrt{\bar{\alpha}_\tau}\textbf{x}_0+\sqrt{1-\bar{\alpha}_\tau}\boldsymbol{\epsilon}_\tau.
   \label{eq: x_t}
\end{equation}

\subsubsection{Reverse Diffusion Process}
The reverse diffusion process progressively removes noise to reconstruct the original data from the noisy state $\textbf{x}_{T}\sim\mathcal{N}(\mathbf{0},\mathbf{I})$.
Since $q(\textbf{x}_{\tau-1}|\textbf{x}_\tau)$ is intractable, this requires a parameterized neural network $p_{\bm{\phi}}(\textbf{x}_{\tau-1}|\textbf{x}_\tau)$ to approximate the conditional probabilities $q(\textbf{x}_{\tau-1}|\textbf{x}_\tau)$ by
\begin{equation}
    p_{\bm{\phi}}(\textbf{x}_{\tau-1}|\textbf{x}_\tau)=\mathcal{N}(\textbf{x}_{\tau-1};\boldsymbol{\mu}_{\bm{\phi}}(\textbf{x}_\tau,\tau),\boldsymbol{\Sigma}_{\bm{\phi}}(\textbf{x}_\tau,\tau)),
    \label{eq: reverse}
\end{equation}
where models $\boldsymbol{\mu}_{\bm{\phi}}(\textbf{x}_\tau,\tau)$ predict the mean at step $\tau$ and $\boldsymbol{\Sigma}_{\bm{\phi}}(\textbf{x}_\tau,\tau) =\sigma_\tau^2\mathbf{I}$. $\sigma_\tau$ can be calculated through $\sigma_\tau^2=\beta_\tau$ or $\sigma_\tau^2=\tilde{\beta}_\tau=\frac{1-\bar{\alpha}_{\tau-1}}{1-\bar{\alpha}_\tau}\beta_\tau$.
$\boldsymbol{\mu}_{\bm{\phi}}(\textbf{x}_\tau,\tau)$ can be parameterized by network $\boldsymbol{\epsilon}_{\bm{\phi}}(\textbf{x}_\tau,\tau)$ that predicts the added noise $\bm{\epsilon}_\tau$ for each step $\tau$.
Thus, $\boldsymbol{\mu}_{\bm{\phi}}(\textbf{x}_\tau,\tau)$ can be expressed as
\begin{equation}
\label{eq.mean}
\boldsymbol{\mu}_{\bm{\phi}}(\textbf{x}_\tau,\tau)=\frac1{\sqrt{\alpha_\tau}}\left(\textbf{x}_\tau-\frac{1-\alpha_\tau}{\sqrt{1-\bar{\alpha}_\tau}}\boldsymbol{\epsilon}_{\bm{\phi}}(\textbf{x}_\tau,\tau)\right).
\end{equation}

\subsubsection{Training Process}
%
DDPM aims to train $\boldsymbol{\mu}_{\bm{\phi}}$ to predict $\tilde{\boldsymbol{\mu}}_{\tau}(\textbf{x}_{\tau},\textbf{x}_0)=\frac1{\sqrt{\alpha_{\tau}}}\left(\textbf{x}_{\tau}-\frac{1-\alpha_{\tau}}{\sqrt{1-\bar{\alpha}_{\tau}}}\boldsymbol{\epsilon}_{\tau}\right)$ when approximating $p_{\bm{\phi}}(\textbf{x}_{\tau-1}|\textbf{x}_\tau)$ in Eq.~\eqref{eq: reverse}, because $\textbf{x}_{\tau}$ can be derived from Eq.~\eqref{eq: x_t}.
To minimize the difference between $\boldsymbol{\mu}_{\bm{\phi}}(\textbf{x}_\tau,\tau)$ and $\tilde{\boldsymbol{\mu}}_\tau$, the loss function of training Process is written as
\begin{equation}
\mathcal{L}_\tau=\mathbb{E}_{\textbf{x}_0,\boldsymbol{\epsilon_\tau}}\Big[\frac1{2\|\boldsymbol{\Sigma}_\theta(\textbf{x}_\tau,\tau)\|_2^2}\|\boldsymbol{\tilde{\mu}}_\tau(\textbf{x}_\tau,\textbf{x}_0)-\boldsymbol{\mu}_{\bm{\phi}}(\textbf{x}_\tau,\tau)\|^2\Big]+C_1,
\end{equation}
where $C_1 > 0$ is a constant.
$\mathcal{L}_\tau$ is further simplified as
\begin{equation}
\mathcal{L}_\tau=\mathbb{E}_{\textbf{x}_0,\boldsymbol{\epsilon}_\tau}\left[\|\boldsymbol{\epsilon}_\tau-\boldsymbol{\epsilon}_{\bm{\phi}}(\textbf{x}_\tau,\tau)\|^2\right]+C_1.
\label{eq: loss}
\end{equation}

\subsubsection{Sampling Process}
Given the mean $\boldsymbol{\mu}_{\bm{\phi}}(\textbf{x}_\tau,\tau)$ and the variance $\mathbf{\Sigma}_{\bm{\phi}}(\textbf{x}_{\tau},\tau)=\sigma_{\tau}^{2}\mathbf{I}$ of the trained network $\boldsymbol{\epsilon}_{\bm{\phi}}(\textbf{x}_\tau,\tau)$, and considering the noise $\textbf{x}_T\sim\mathcal{N}(\mathbf{0},\mathbf{I})$, for $\tau \in \{T,\ldots,1\}$, $\textbf{x}_{\tau-1}$ is sampled by
\begin{equation}
    \textbf{x}_{\tau-1}=\frac{1}{\sqrt{\alpha_{\tau}}}\left(\textbf{x}_{\tau}-\frac{1-\alpha_{\tau}}{\sqrt{1-\bar{\alpha}_{\tau}}}\boldsymbol{\epsilon}_{\bm{\phi}}(\textbf{x}_{\tau},\tau)\right)+\sigma_{\tau}\mathbf{z},
\end{equation}
where $\mathbf{z}\sim\mathcal{N}(\mathbf{0},\mathbf{I})$ for $\tau>1$, otherwise $\mathbf{z}=\mathbf{0}$.

\begin{figure}[t]
    \centering
    \includegraphics[width=1\linewidth]{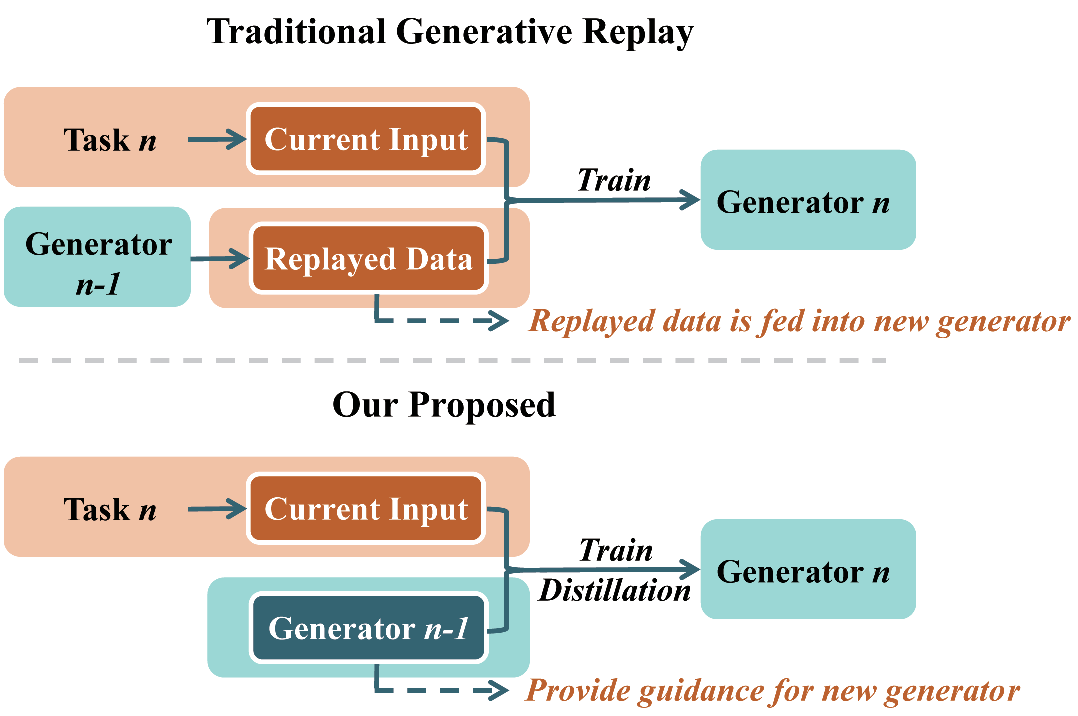}
    \caption{Comparison of traditional generative replay
    and our proposed DSG.
    }
    \label{fig:distillation}
\end{figure}

\subsection{Distillation-based Self-Guidance}
As illustrated in Fig.~\ref{fig:distillation}, traditional generative replay typically updates the generator using replay data generated from previous versions of the generator itself.
Our DSG enhances this process by directly employing the previous generator to train the new generator, thereby providing a more direct form of guidance.
We summarize the CL process in Algorithm~\ref{alg:dsg}.

We set the same value of $T$ for all tasks.
The self-guidance loss during the generator update stage consists of two components: one for the current specific tasks $\mathcal{L}_{task}$ and another for distillation $\mathcal{L}_{dist}$.
When the generator is learning the input from current task $\mathcal{T}_n$,
$\mathcal{L}_{n, task}$ is calculated by
\begin{equation}
    \mathcal{L}_{n,task}= \mathbb{E}_{\tau \sim  \left [ 1, T\right ], \textbf{x}_{n}, \bm{\epsilon}_{\bm{\phi}_{n}} }\left[\left\|\boldsymbol{\epsilon}_{n,\tau}-\boldsymbol{\epsilon}_{\bm{\phi}_{n}}\left(\textbf{x}_{n,\tau}, \tau\right)\right\|^{2}\right] + C_2,
\end{equation}
where $\textbf{x}_{n,\tau}$ is computed as $\sqrt{\bar{\alpha}_\tau}\textbf{x}_{n,0}+\sqrt{1-\bar{\alpha}_\tau}\bm{\epsilon}$ (by Eq.~\eqref{eq: x_t}), $\bm{\phi}_{n}$ represents the parameter of updated generator following the learning of task $\mathcal{T}_n$, and $C_2 > 0$ denotes a constant.

The distillation loss aims to ensure that each addition of Gaussian noise by the new generator closely resembles the addition of noise by the old generator during the forward process.
When the generator is learning task $\mathcal{T}_n$, its distillation loss $\mathcal{L}_{n,dist}$ is expressed as
\begin{equation}
\begin{split}
\mathcal{L}_{n,dist} & = \mathbb{E}_{\tau \sim  \left [ 1, T\right ], \textbf{x}_{n}, \bm{\epsilon}_{\bm{\phi}_{n}} } \\
& \quad \left[\left\|\boldsymbol{\epsilon}_{\bm{\phi}_{n-1}}\left(\textbf{x}_{n,\tau}, \tau\right)-\boldsymbol{\bm{\epsilon}}_{\bm{\phi}_n}\left(\textbf{x}_{n,\tau}, \tau\right)\right\|^{2}\right] + C_3,
\end{split}
\end{equation}
where $\bm{\phi}_{n-1}$
represents the parameter of old generator which is updated following the learning of task $n-1$ and $C_3 > 0$ is a constant.

Combining the new task-specific loss and the distillation loss together, the overall loss function when the generator is learning task $n$ is
\begin{equation}
\label{dsg_loss}
\begin{split}
\mathcal{L}_{n} & = \mathcal{L}_{n,task} + \lambda \mathcal{L}_{n,dist} \\
& = \mathbb{E}_{\tau \sim  \left [ 1, T\right ], \textbf{x}_{n}, \bm{\epsilon}_{\bm{\phi}_{n}} } \left[\left\|\boldsymbol{\epsilon}_{n,\tau}-\boldsymbol{\epsilon}_{\bm{\phi}_{n}}\left(\textbf{x}_{n,\tau}, \tau\right)\right\|^{2}  \right.\\
& \quad \left. + \lambda \left\|\boldsymbol{\epsilon}_{\bm{\phi}_{n-1}}\left(\textbf{x}_{n,\tau}, \tau\right)-\boldsymbol{\epsilon}_{\bm{\phi}_{n}}\left(\textbf{x}_{n,\tau}, \tau\right)\right\|^{2}\right ] + C_4,
\end{split}
\end{equation}
where $\lambda$ is a constant determining the weight of distillation and $C_4 > 0$ is a constant.

\begin{algorithm}[t]
\DontPrintSemicolon
\caption{Generative-based Replay CL with DSG}
\label{alg:dsg}
\KwIn{Sequence of tasks $\left \{  \mathcal{T}_{1}, \mathcal{T}_{2}, ..., \mathcal{T}_{N}\right \}$}
\KwOut{Continual learned model $M_{\bm{\theta}_N}$, Generator $G_{\bm{\phi}_N}$}
\tcp{Initialization}
$M_{\bm{\theta}_1}, G_{\bm{\phi}_1} \leftarrow \mathcal{T}_{1}$\;
\tcp{Continual Training}
\For{$\mathcal{T}_{n}\in(\mathcal{T}_{2}, \cdots, \mathcal{T}_{N}$)}  {
    \tcc{Model update}
    Generate replay samples $\textbf{x}_n'$ using $G_{\bm{\phi}_{n-1}}$\;
    Label the replay samples $y_n'$ using $M_{\bm{\theta}_{n-1}}$\;
    Update the current model $M_{\bm{\theta}_n}$ using Stochastic Gradient Descent by learning from $\mathcal{T}_{n}$ and $(\textbf{x}_n', y_n')$\;
    \tcc{Generator Update with DSG}
    Update the current generator using $\mathcal{T}_{n}$ by Eq.~\eqref{dsg_loss}\;
}
\end{algorithm}

\section{Experiments}
\subsection{Datasets}
We conduct experiments on three industrial datasets: CWRU \cite{smith2015rolling}, DSA \cite{altun2010human}, and WISDM \cite{weiss2019wisdm}, which are described in Table~\ref{tab:datasets}.

1) \textbf{CWRU} dataset \cite{smith2015rolling} comprises vibration signal data collected under various operating conditions involving different fault types such as inner race, outer race, and rolling elements. Fault sizes range from 0.007 inches to 0.021 inches.

2) \textbf{DSA} dataset \cite{altun2010human} comprises data from 19 daily activities conducted by 8 participants. It covers 45 channels over 125 time steps, with 18 activity categories selected for analysis to ensure analytical balance. \looseness=-1

3) \textbf{WISDM} dataset \cite{weiss2019wisdm} comprises monitoring and classification data of sensor-based human activities, covering 18 different activities performed by 51 participants. Data is collected via smartphone accelerometers and segmented into non-overlapping sliding windows of 200 data points. Each sample represents a 10-second time sequence, sampled at a rate of 20 Hz.

\begin{table}[t]
\centering
\caption{Overview of the selected datasets.}
\label{tab:datasets}
\begin{tabular}{@{}lccc@{}}
\toprule
\textbf{Datasets} & \textbf{Shape (C$\times$L)} & \textbf{Train/Test Size} & \textbf{Classes/Exp Tasks} \\ \midrule
CWRU          & 2$\times$1024                 & 22400/5600               & 10/5                     \\
DSA              & 45$\times$125                & 6840/2280               & 18/6                     \\
WISDM            & 3$\times$200                 & 18184/6062               & 18/6                     \\ \bottomrule
\end{tabular}
\end{table}

\begin{figure*}[!t]
  \centering
  \begin{subfigure}[b]{0.3\linewidth}
    \includegraphics[width=\linewidth]{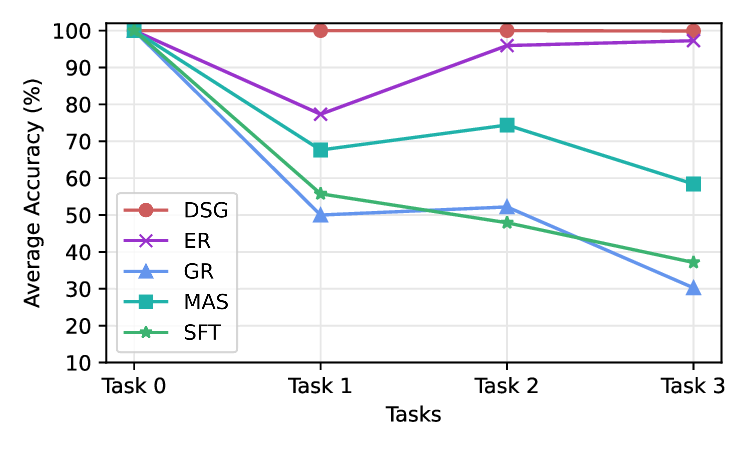}
    \caption{CWRU.}
    \label{fig:1}
  \end{subfigure}
  \begin{subfigure}[b]{0.3\linewidth}
    \includegraphics[width=\linewidth]{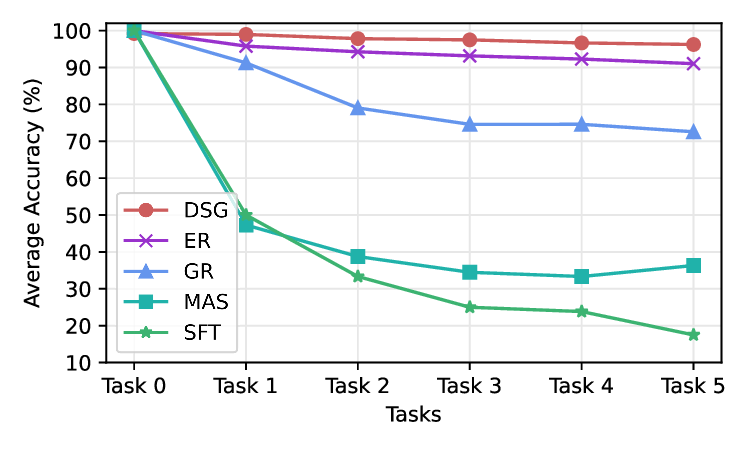}
    \caption{DSA.}
    \label{fig:2}
  \end{subfigure}
  \begin{subfigure}[b]{0.3\linewidth}
    \includegraphics[width=\linewidth]{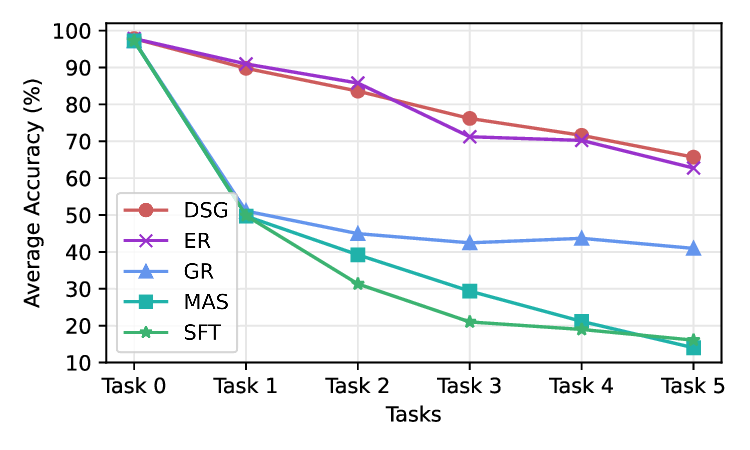}
    \caption{WISDM.}
    \label{fig:3}
  \end{subfigure}
\caption{Evolution of average accuracy of various methods.}
\end{figure*}

\subsection{Comparison Methods}
We compare DSG with the following baseline methods:

1) \textbf{SFT} \cite{ohri2024supervised} sets the lower performance bound by sequentially fine-tuning the model on tasks without employing CL techniques.

2) \textbf{MAS} \cite{aljundi2018memory} quantifies parameter importance based on their sensitivity to the predicted output function, retaining significant weights for previous tasks.

3) \textbf{GR} \cite{shin2017continual} uses GANs for generative replay to prevent forgetting and simultaneously reduce the storage burden.

4) \textbf{ER} \cite{rolnick2019experience} stores a portion of data samples from previous tasks in a memory buffer. These stored data are then replayed for training when the continual model learns a new task.

\subsection{Evaluation Metrics}
1) \textbf{Average Accuracy}:
The average accuracy is measured after the completion of a CL at task $n$ by
\begin{equation}
    A_n = \frac{1}{n}\sum_{j=1}^{n}a_{n,j},
\end{equation}
where $a_{n,j}\in[0,1]$ represents the classification accuracy evaluated on the test set of the $j$-th task after the continual model has learned the $n$-th task.

2) \textbf{Forgetting Measure}:
For a continual model that has learned $n$ tasks, task forgetting for the previous $j$-th task $(j\leq n)$ is calculated as the difference between its maximum previous performance and its current performance:
\begin{equation}
    f_{j}^{n}=\max_{i\in\{1,\ldots,n-1\}}(a_{i,j}-a_{n,j}),\forall j<n.
\end{equation}
The average forgetting at the $n$-th task is then calculated as
\begin{equation}
    \mathrm{F}_n=\frac1{n-1}\sum_{j=1}^{n-1}f_{j}^{n}.
\end{equation}

\subsection{Model Architecture}
The classifier design of the model used is described in detail in Table~\ref{tab:network_architecture}.
The model is a CNN encoder explicitly designed for processing one-dimensional input signals.
It comprises multiple convolutional blocks, each consisting of a convolutional layer with ReLU activation and batch normalization, followed by max pooling and dropout for regularization.
The network progressively increases the feature dimensionality as it deepens, ultimately leading to a global average pooling layer to generate the final feature representation.
Inspired by insights from Desai \textit{et al.} \cite{desai2021timevae}, our diffusion probabilistic model integrates the Unet-1D framework, a VAE architecture designed to generate multivariate time series data.

\begin{table}[t]
\centering
\caption{Detailed configuration of the CNN encoder.}
\label{tab:network_architecture}
\begin{tabular}{@{}lccl@{}}
\toprule
\textbf{Layer} & \textbf{Kernel Size} & \textbf{Channels} & \textbf{Other Parameters} \\ \midrule
Conv1d-1       & 5                    & 64                & Stride=1, Padding=2       \\
Conv1d-2       & 5                    & 128               & Stride=1, Padding=2       \\
Conv1d-3       & 5                    & 256               & Stride=1, Padding=2       \\
Conv1d-4       & 5                    & 128               & Stride=1, Padding=2       \\
Global Avg Pool & 2                   & -                  & Stride=1, Padding=0                         \\ \bottomrule
\end{tabular}
\end{table}

\subsection{Implementation Details}
An early stopping strategy with a patience value of 20 is employed to prevent model overfitting.
For each task, the ratio of the training set to the validation set is 9:1.
Early stopping is activated when the loss on the validation set ceases to decrease.
When adjusting hyperparameters, the optimal learning rate is 0.001, and the batch size is set to 64.
The dropout rate is set to 0.1 for the CWRU dataset and 0.3 for the others.
The batch size for training and the replay buffer is standardized at 32, ensuring a consistent number of samples used at each training step.
Since experiments are conducted five times with different class sequences, optimal hyperparameter configurations may vary between runs. Configurations are adjusted based on the specific conditions of each experiment.

\subsection{Experiments Analysis}

1) \textbf{Average Accuracy}:
The results for all three datasets are presented in Table~\ref{tab:results}. The evolution of average accuracy across datasets is illustrated in Figs.~\ref{fig:1} to \ref{fig:3}.
DSG demonstrates improvements of 2.9\%, 5.0\%, and 3.0\% in accuracy compared to the optimal baseline ER on the CWRU, DSA, and WIDM datasets, respectively.
GR, ER, and DSG achieve higher accuracy on the CWRU and DSA datasets but underperform on the WISDM dataset.
This performance discrepancy is likely due to the differing complexities among the datasets. The complexities are illustrated in Table~\ref{tab:datasets}, where the data samples exhibit distinct characteristics.
This limitation hampers the balanced assessment of previous categories.
DSG significantly improves performance in these areas by introducing a distillation-guided generator, effectively addressing the issues above.

\begin{table*}[t]
\centering
\caption{Performances on three datasets.}
\label{tab:results}
\begin{tabular}{lllllll}
\toprule
\multicolumn{1}{c}{} & \multicolumn{3}{c}{Final Average Accuracy ($\uparrow$)}                                         & \multicolumn{3}{c}{Forgetting Measure ($\downarrow$)}                                            \\
\cline{2-7}
\multicolumn{1}{c}{} & \multicolumn{1}{c}{CWRU} & \multicolumn{1}{c}{DSA} & \multicolumn{1}{c}{WIDM} & \multicolumn{1}{c}{CWRU} & \multicolumn{1}{c}{DSA} & \multicolumn{1}{c}{WISDM} \\ \toprule
SFT                  & 25.1\tiny($\pm$ 5.7)                 & 16.9\tiny($\pm$ 2.5)                & 16.1\tiny($\pm$ 4.7)                 & 97.2\tiny($\pm$ 3.6)                       & 96.1\tiny($\pm$ 3.1)                      & 95.6\tiny($\pm$ 3.3)                        \\
MAS                  & 59.3\tiny($\pm$ 4.9)                 & 37.1\tiny($\pm$ 4.5)                & 14.0\tiny($\pm$ 3.9)                 & 49.3\tiny($\pm$ 15.1)                       & 55.1\tiny($\pm$ 10.7)                      & 66.2\tiny($\pm$ 5.4)                        \\
GR                   & 31.0\tiny($\pm$ 5.0)                 & 72.6\tiny($\pm$ 5.7)                & 41.5\tiny($\pm$ 5.0)                 & 37.7\tiny($\pm$ 6.9)                       & 33.2\tiny($\pm$ 4.5)                      & 45.7\tiny($\pm$ 6.1)                        \\
ER                   & 96.8\tiny($\pm$ 2.6)                 & 91.0\tiny($\pm$ 3.5)                & 62.7\tiny($\pm$ 2.6)                 & \textbf{24.3\tiny($\pm$ 2.4)}                       & 22.3\tiny($\pm$ 4.9)                      & \textbf{25.9\tiny($\pm$ 8.1)}                        \\
DSG                  & \textbf{99.7\tiny($\pm$ 0.9)}                 & \textbf{96.0\tiny($\pm$ 1.0)}                & \textbf{65.7\tiny($\pm$ 1.9)}                 & 33.6\tiny($\pm$ 3.4)                       & \textbf{21.9\tiny($\pm$ 5.1)}                      & 35.5\tiny($\pm$ 14.1)                        \\ \bottomrule
\end{tabular}
\end{table*}

2) \textbf{Forgetting Measure}:
DSG does not consistently outperform the ER approach regarding the forgetting rate.
The forgetting rate increases by 9.3\% and 9.6\% relative to the ER on the CWRU and WISDM datasets, respectively.

\begin{figure}[t]
    \centering
    \includegraphics[width=0.98\linewidth]{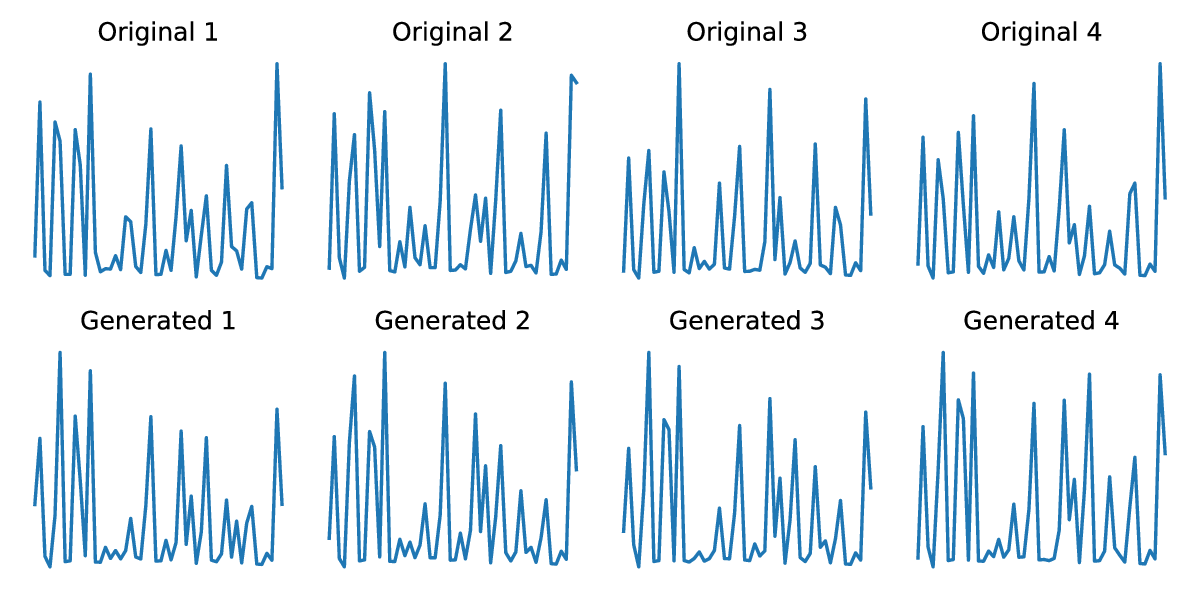}
    \caption{
    Comparison of raw and generative samples.
    }
    \label{fig:comparison}
\end{figure}

3) \textbf{Visualization of Synthesized Samples}:
We present several original and generated samples from the CWRU dataset in Fig.~\ref{fig:comparison}.
For ease of observation, we display the first 50 data points of one dimension of each sample.
The generated samples exhibit similarity to the original samples. This suggests that DSG can synthesize real patterns in industrial streaming data. \looseness=-1

\begin{figure*}[t]
    \centering
    \begin{subfigure}{0.25\textwidth}
        \includegraphics[width=\linewidth]{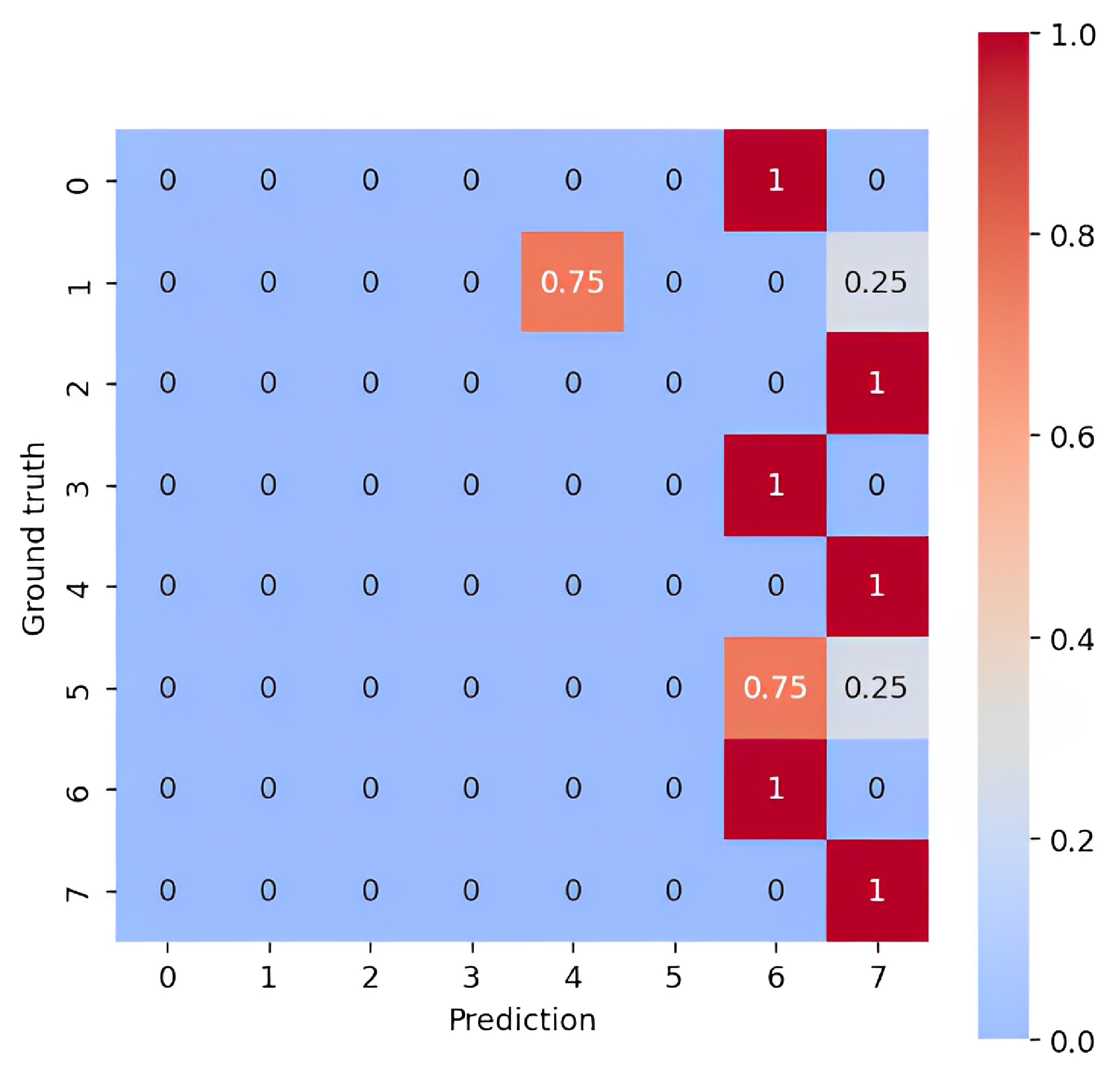}
        \caption{SFT}
        \label{fig:sub1}
    \end{subfigure}
    \begin{subfigure}{0.25\textwidth}
        \includegraphics[width=\linewidth]{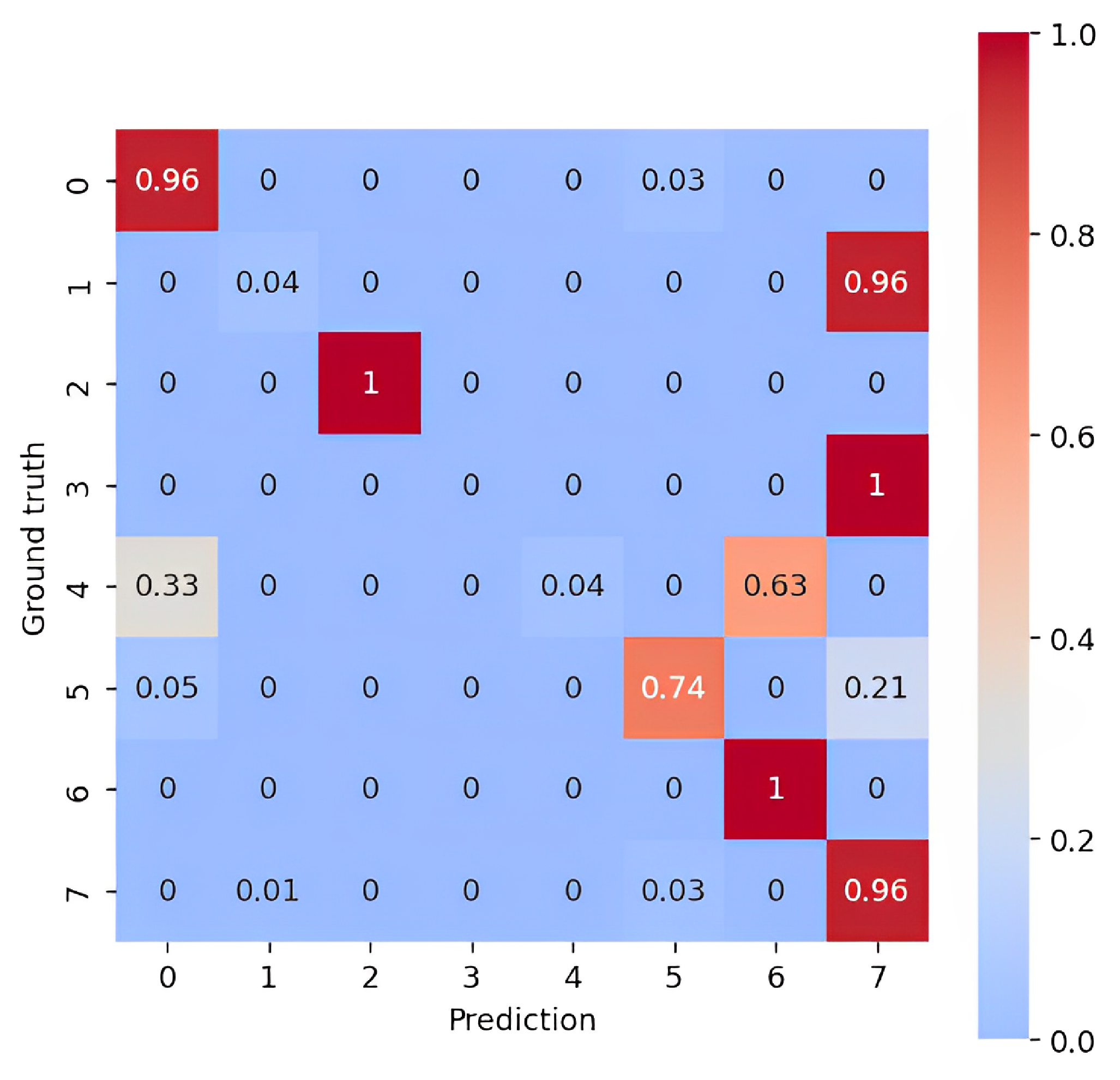}
        \caption{MAS}
        \label{fig:sub2}
    \end{subfigure}
    \begin{subfigure}{0.25\textwidth}
        \includegraphics[width=\linewidth]{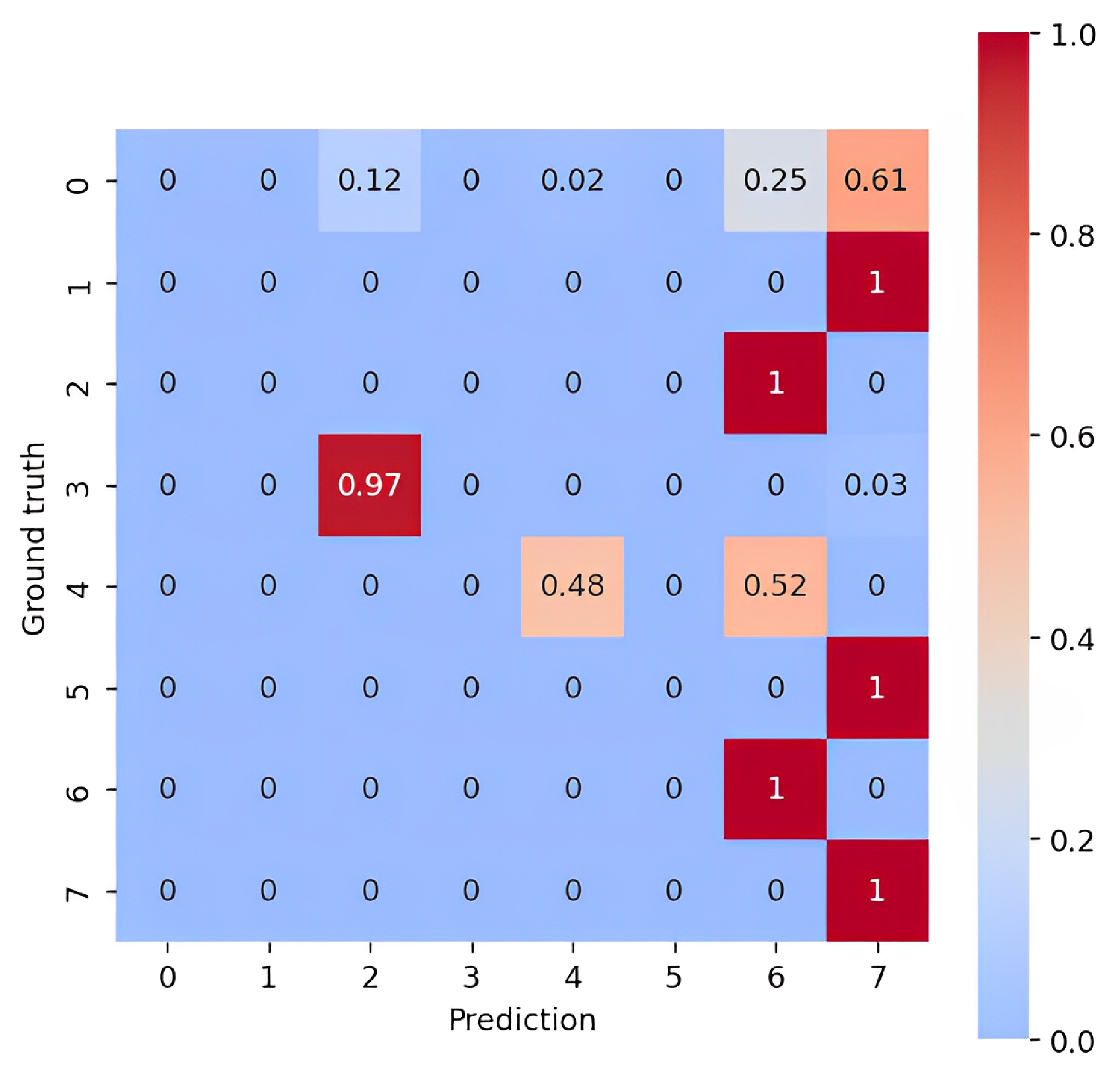}
        \caption{GR}
        \label{fig:sub3}
    \end{subfigure}
    \begin{subfigure}{0.25\textwidth}
        \includegraphics[width=\linewidth]{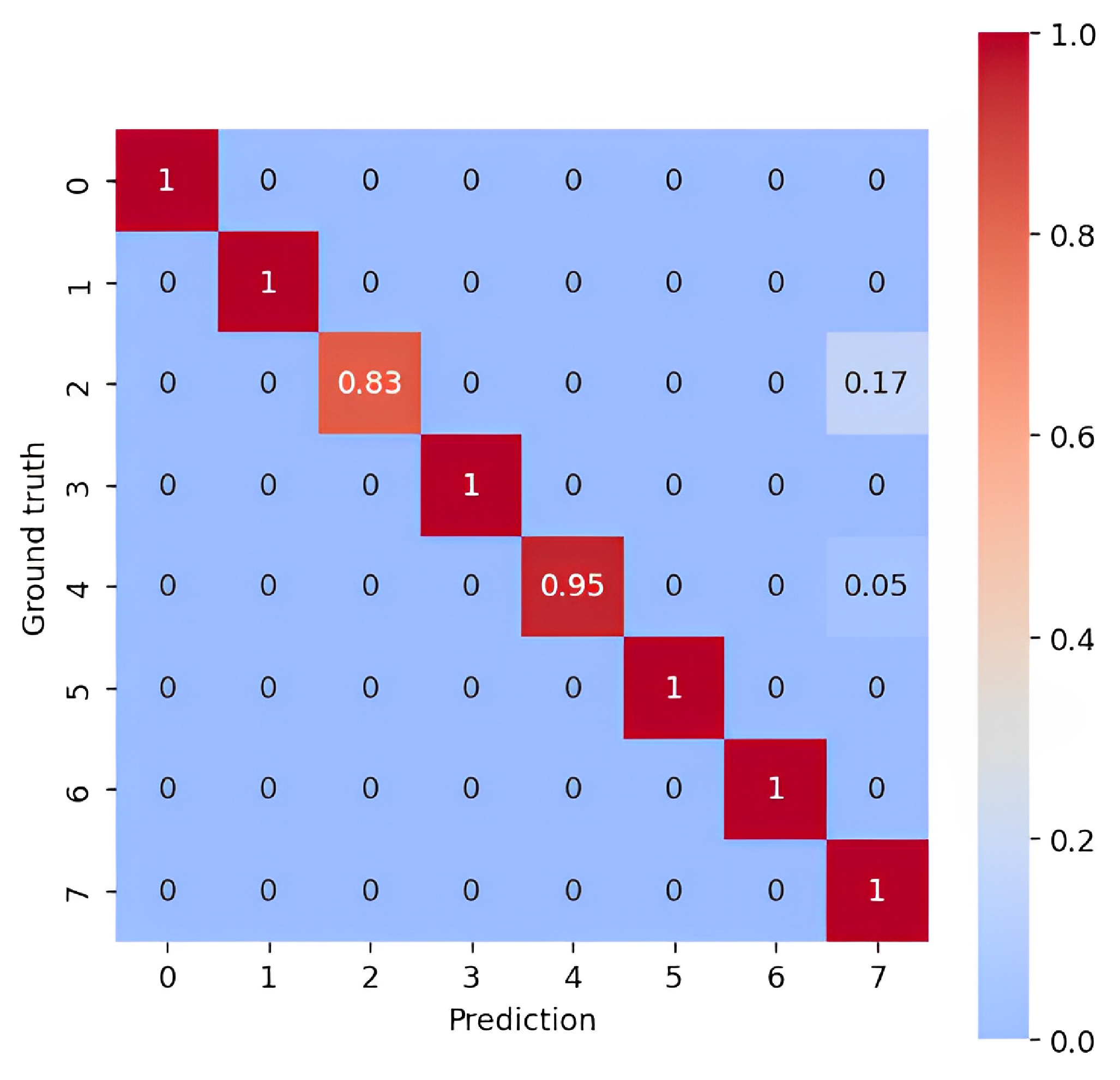}
        \caption{ER}
        \label{fig:sub4}
    \end{subfigure}
    \begin{subfigure}{0.25\textwidth}
        \includegraphics[width=\linewidth]{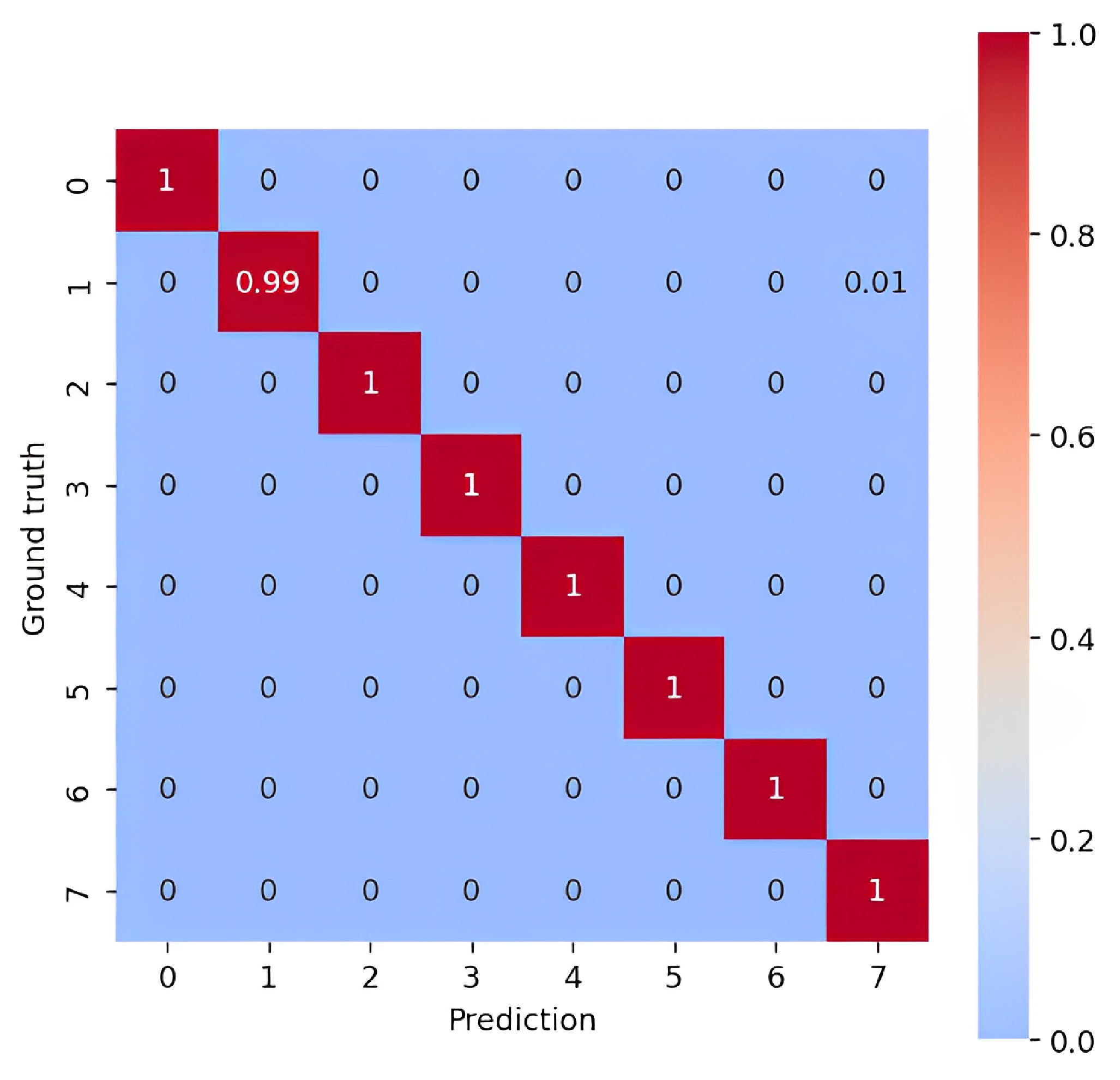}
        \caption{DSG}
        \label{fig:sub5}
    \end{subfigure}
    \caption{Comparing DSG with the baselines among continual tasks using the CWRU dataset.}
    \label{fig:test}
\end{figure*}

4) \textbf{Confusion Matrices}:
The confusion matrices for all methods on the CWRU dataset are shown in Fig.~\ref{fig:sub1} to \ref{fig:sub5}.
While all methods achieve high accuracy in recognizing the two most recent categories (6 and 7), the SFT, MAS, and GR methods encounter difficulties in differentiating previously learned categories (0 to 5).
ER and DSG demonstrate stronger robustness in resisting forgetting.

\section{Conclusion}
This paper explored CL of streaming data for the IIoT, where the challenge is to adapt to new data distributions while mitigating catastrophic forgetting in resource-constrained devices.
We proposed a novel generative replay CL approach that utilizes Distillation-based Self-Guided to achieve efficient knowledge transfer from old generators to new ones, significantly improving the quality of replay data.
We evaluated our approach on the CWRU, DSA, WISDM datasets and showed that the proposed DSG enhances the model's adaptability while effectively mitigating catastrophic forgetting.
A direction of future work involves exploring methods to extract more knowledge from new tasks' data and replayed data in IIoT scenarios from a continual model perspective.
The model can be deployed through cloud and local/edge collaboration for cost savings and data security.

\bibliographystyle{IEEEtran}
\bibliography{IEEEabrv,references}
\end{document}